\documentclass[letterpaper]{article} 
\usepackage{aaai2026}  
\usepackage{times}  
\usepackage{helvet}  
\usepackage{courier}  
\usepackage[hyphens]{url}  
\usepackage{graphicx} 
\usepackage{natbib}  
\usepackage{caption} 
\usepackage{algorithm}
\usepackage{algorithmic}
\usepackage{times}  
\usepackage{helvet}  
\usepackage{courier}  
\usepackage[hyphens]{url}  
\usepackage{graphicx} 
\usepackage{natbib}  
\usepackage{caption} 
\usepackage[table,xcdraw]{xcolor}
\usepackage{algorithm}
\usepackage{algorithmic}
\usepackage{booktabs}
\usepackage{graphicx}	
\usepackage{amsmath}
\usepackage{amssymb}	
\usepackage{booktabs}
\usepackage{microtype}
\usepackage{epsfig}
\usepackage{float}
\usepackage{placeins}
\usepackage{color, colortbl}
\usepackage{enumitem}
\usepackage{tabularx}
\usepackage{xstring}
\usepackage{multirow}
\usepackage{xspace}
\usepackage{subcaption}
\usepackage{xcolor}
\definecolor{brownish-red}{RGB}{165,42,42}
\definecolor{FireBrick-red}{RGB}{178,34,34}
\usepackage{bibentry}
\usepackage{newfloat}
\usepackage{listings}
\DeclareCaptionStyle{ruled}{labelfont=normalfont,labelsep=colon,strut=off} 
\floatstyle{ruled}
\newfloat{listing}{tb}{lst}{}
\floatname{listing}{Listing}
\title{A Closer Look at Knowledge Distillation in Spiking Neural Network Training}
\author{
    Xu Liu\textsuperscript{\rm 1},
    Na Xia\textsuperscript{\rm 1}\thanks{Corresponding authors},
    Jinxing Zhou\textsuperscript{\rm 3},
    Jingyuan Xu\textsuperscript{\rm 1},
    Dan Guo\textsuperscript{\rm 1,2*}
}
\affiliations{
    \textsuperscript{\rm 1}School of Computer Science and Information Engineering, Hefei University of Technology\\
    \textsuperscript{\rm 2}Institute of Artificial Intelligence, Hefei Comprehensive National Science Center\\
    \textsuperscript{\rm 3}Mohamed Bin Zayed University of Artificial Intelligence\\
}

\begin{document}

\maketitle

\begin{abstract}
Spiking Neural Networks (SNNs) become popular due to excellent energy efficiency, yet facing challenges for effective model training.
Recent works improve this by introducing knowledge distillation (KD) techniques, with the pre-trained artificial neural networks (ANNs) used as teachers and the target SNNs as students.
This is commonly accomplished through a straightforward element-wise alignment of intermediate features and prediction logits from ANNs and SNNs, often neglecting the intrinsic differences between their architectures. 
Specifically, ANN's outputs exhibit a continuous distribution, whereas SNN's outputs are characterized by sparsity and discreteness.
To mitigate this issue, we introduce two innovative KD strategies. 
Firstly, we propose the 
Saliency-scaled 
Activation Map Distillation (SAMD), which aligns the spike activation map of the student SNN with the class-aware activation map of the teacher ANN. Rather than performing KD directly on the raw 
features of ANN and SNN, our SAMD directs the student to learn from saliency activation maps that exhibit greater semantic and distribution consistency.
Additionally, we propose a Noise-smoothed Logits Distillation (NLD), which utilizes Gaussian noise to smooth the sparse logits of student SNN, facilitating the alignment with continuous logits from teacher ANN.
Extensive experiments on multiple datasets demonstrate the effectiveness of our methods.
Code is available~\footnote{\url{https://github.com/SinoLeu/CKDSNN.git}}. 
\end{abstract}


\section{Introduction}
\label{sec:intro}
Spiking Neural Networks (SNNs), 
inspired by the spiking mechanism of biological neurons, utilize event-driven binary spikes to transmit information,
allowing multiplications between activations and weights to be replaced by additions or remain silent, thereby significantly improving energy efficiency~\cite{eshraghian2021training, davies2018loihi}. 
Taking advantage of this computational paradigm, SNNs can operate efficiently on neuromorphic hardware and demonstrate autonomous learning capabilities and ultralow power consumption~\cite{mehonic2022brain}, making them highly promising for intelligent computing tasks~\cite{fang2023spikingjelly, zhang2020system}.
However, training SNNs presents significant challenges due to the 
inherently discrete and sparse nature of spike-based features, 
which complicates their optimization process and results in performance and application limitations compared to traditional artificial neural networks (ANNs)~\cite{zhou2021positive,zhou2022audio,zhou2023contrastive,zhou2024avss,zhou2024towards,zhou2024label,zhou2024dense,zhou2024vaplan,li2025patch,li2024object,qian2025physdiff,zhao2025multimodal,zhou2025mettle,zhou2025think,jin2025simtoken,zhou2025clasp,kryklyvets2025mavis,shen2023fine,song2022memorial,guo2025audio}.

\begin{figure}[tp]
    \centering
    \includegraphics[width=1\linewidth]{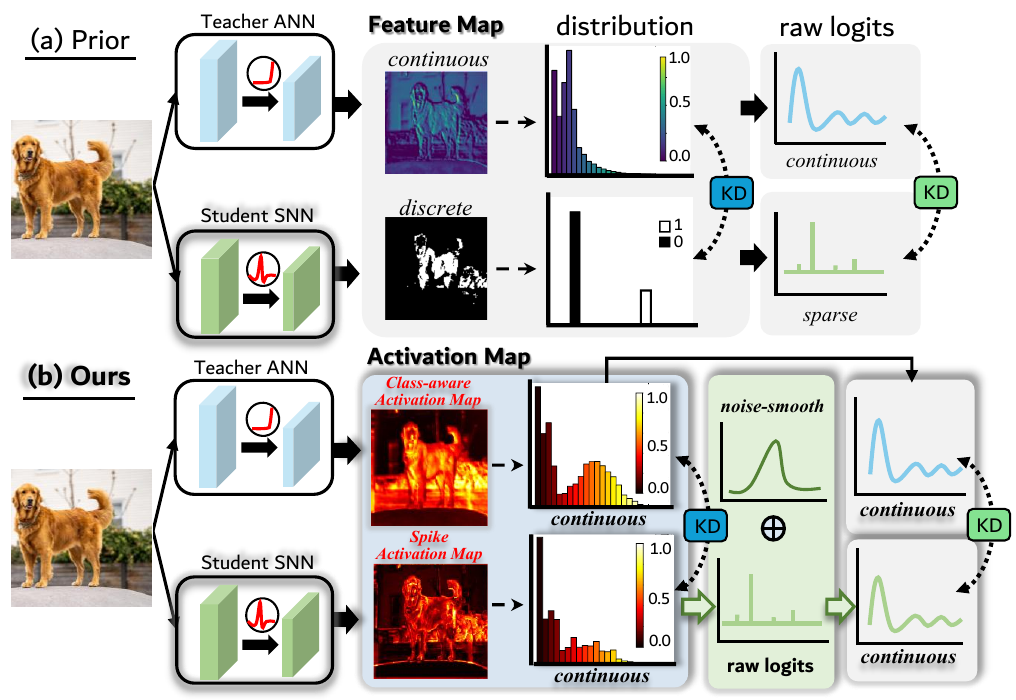}
    \vspace{-4ex}
    \caption{(a) Prior KD methods simply aligns the raw hidden features and output logits between teacher ANN and student SNN, ignoring discrepancies in their distributions.
    (b) We perform the KD through more precise and semantic-consistent saliency maps, aligning the spiking activation map of SNN with the class activation map with ANN.
    Besides, we utilize Gaussian noise to smooth the raw logits of SNN, reducing the discrepancy in logits distillation.
    }\label{fig1}
    \vspace{-2ex}
\end{figure}

Specifically, conversion-based methods~\cite{bu2022optimal} transfer pre-trained ANN parameters to corresponding SNNs but replace ReLU activation with spiking neurons~\cite{huang2024clif}. 
This strategy has been experimentally found to require a large number of time steps to achieve satisfactory performance~\cite{bu2022optimal,han2020rmpsnn,Rueckauer2017conversion}. 
Direct training methods~\cite{wu2019direct,fang2021sew,deng2022tet}, on the other hand, optimize SNNs using direct backpropagation through the surrogate gradient estimation technique~\cite{zenke2021remarkable}, leading to significant progress in pattern recognition~\cite{deng2022tet,zhou2023spikformer}, natural language processing
~\cite{zhang2024spikingminilm,xiao2022towards}, and multimodal tasks~\cite{LiuMISNet2025}.
While this strategy reduces training time steps, 
the SNN's performance still lags behind that of ANNs.

Recent works~\cite{xu2025bkdsnn,Xu2023kdsnn} improve SNN training using knowledge distillation (KD)~\cite{hinton2014distilling} techniques, with pretrained ANNs as the teacher model and SNNs as the student model, showing promising results on multiple datasets.
These methods perform KD by element-wise aligning the output features~\cite{Xu2023kdsnn} or classification logits~\cite{Xu2023kdsnn} between ANNs and SNNs. 
Such a KD paradigm distills knowledge from teacher ANNs to student SNNs, however, prior works ignore two critical issues: 
\textbf{(i) Discrepancy between raw features.}
As illustrated in Fig.~\ref{fig1}(a), 
features extracted from ANNs  via a single forward pass are represented as \textit{continuous floating-point} values. In contrast, features in SNNs, obtained by forward propagation over multiple time steps, are expressed as \textit{discrete binary spikes}. Moreover, whereas ANN features encapsulate patterns spanning the entire image, SNN spikes primarily highlight \textit{salient regions}.
\textbf{(ii) Discrepancy between raw logits.} The logits (\textit{i.e.}, raw classification scores) are derived from the hidden features. Consequently, a notable disparity emerges between the raw logits of the two models. Specifically, SNN logits display greater sparsity and a more peaked distribution relative to those of ANNs.

In this paper, 
we propose novel knowledge distillation strategies, 
having a Closer look at {KD} for {SNN}.
Specifically, to address the first challenge, we propose the \textbf{Saliency-scaled Activation Map Distillation (SAMD)}.
As illustrated in Fig.~\ref{fig1}, unlike previous methods that directly perform knowledge distillation using raw features, 
our SAMD leverages the Class Activation Map (CAM)~\cite{zhou2016learning} of the teacher ANN, which provides more precise and focused knowledge, clearly describing the salient image regions related to the target class. 
Notably, unlike traditional activation map-based distillation methods from ANNs (\textit{e.g.}, e$^2$KD~\cite{parchami2024good} and CATKD~\cite{guo2023class}), we discover that the surrogate gradient estimation in SNNs prevents the use of precise gradient estimation methods like Grad-CAM~\cite{selvaraju2017grad} to generate saliency activation maps.
Instead, we redesign the activation map distillation by aligning the Spiking Activation Map (SAM) of the student SNN with the CAM of the teacher ANN.
Although both SAM and CAM originate from features, they are more consistent than raw features due to the use of saliency maps.
Furthermore, we consider the numerical magnitude differences between the activation maps generated from SNN features (\textit{i.e.}, SAM) and ANN features (\textit{i.e.}, CAM).
To more accurately assess the contribution of each pixel in the saliency maps, we apply the softmax function to convert both CAM and SAM into probability distributions.
In this way, the scaled CAM and SAM remain consistent in both semantics and numerical magnitude, facilitating their alignment.
To address the second challenge, we propose the \textbf{Noise-smoothed Logits Distillation (NLD)}.
As demonstrated in Fig.~\ref{fig1}(b), NLD employs Gaussian noise to moderate the prediction logits of student SNNs.
Specifically, we sample Gaussian noise with mean and variance parameters derived from the SNN logits, ensuring that the original distribution of these logits remains largely preserved.
After the addition of noise, the logits of SNNs transit from a sparse and sharply peaked distribution to one that is denser and broader, resembling the distribution of ANN logits, thereby facilitating knowledge transmission between teacher and student.
We evaluate the effectiveness and superiority of the proposed two KD strategies on CIFAR-10, CIFAR-100, and ImageNet-1K da of SNNs.

In summary, our main contributions are as follows:
\begin{itemize}
    \item We propose a saliency-scaled activation map distillation strategy that directs the student SNN's spike activation map to align with the teacher ANN's class activation map, emphasizing spike generation in salient image regions to improve knowledge transfer. 
    \item We propose a noise-smoothed logits distillation strategy that employs Gaussian noise to moderate the sparse logits of the student SNN, facilitating alignment with the continuous logits of the teacher ANN.
    \item Our method achieves new state-of-the-art performance on multiple datasets and can be flexibly integrated into existing KD approaches for SNN training, maintaining a good balance between accuracy and energy efficiency.
\end{itemize}
\section{Related Work}
\label{sec:related} 

\noindent  \textbf{Spiking Neural Networks (SNNs)} are brain-inspired models that mimic biological neural systems by transmitting information through discrete spikes, 
achieving lower energy consumption. SNNs are typically trained using two main approaches: 
ANN-SNN conversion~\cite{meng2022training,bu2022optimal,deng2021optimal,hu2023spiking} 
and direct training~ \cite{fang2021sew,guo2023rmp,guo2023mbpn,deng2022tet,meng2023towards}.
The conversion methods directly transform pre-trained ANN into SNN by replacing its ReLU activation functions with 
integrate-and-fire (IF)~\cite{bu2022optimal} neurons.
However, the converted SNNs often require prolonged time steps to collect sufficient spike signals to ensure accuracy.
The direct training methods directly train the SNN by backpropagating surrogate gradients~\cite{fang2021sew} through multiple time steps,
which can alleviate excessive time steps, enabling efficient training and inference. However, a significant accuracy gap persists between the trained SNNs and ANNs due to the approximation errors~\cite{Xu2023kdsnn} inherent in surrogate gradients. 
Unlike them, our method utilizes the informative knowledge in pretrained ANNs to better supervise SNN training, achieving balance between accuracy and efficiency.

\begin{figure*}
    \centering
    \includegraphics[width=1\linewidth]{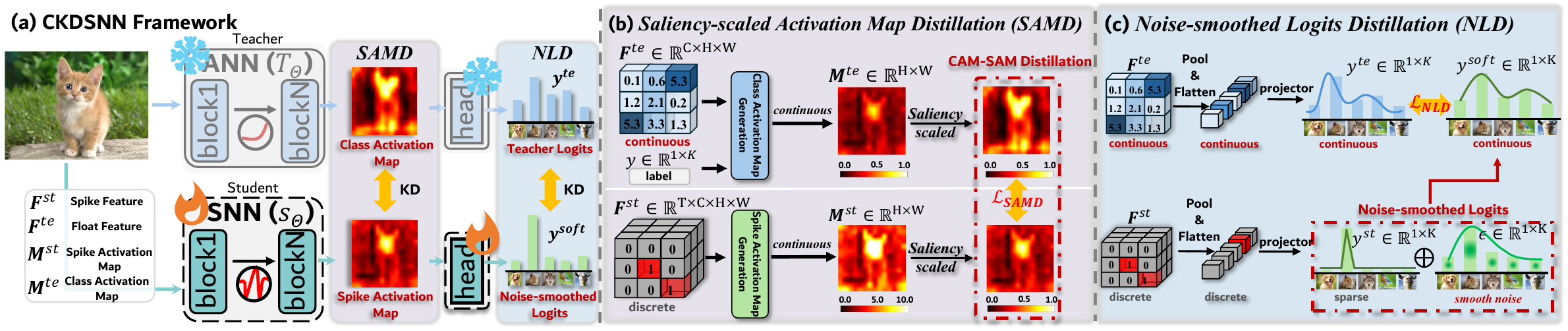}
    \vspace{-4ex}
    \caption{Overview of our CKDSNN.
    (a) CKDSNN framework aims to improve the student SNN training by distilling knowledge from a pretrained teacher ANN.
    CKDSNN is primarily composed of two strategies.
    (b) The \textit{\textbf{Saliency-scaled Activation Map Distillation (SAMD)}} utilizes the class activation map (CAM) from the ANN to guide the SNN to generate precise spike activations in salient regions, \textit{i.e.}, the spike activation map (SAM). Saliency-scaled is used to scale the CAM and SAM into magnitude-unified distributions. 
    (c) The \textit{\textbf{Noise-smoothed Logits Distillation (NLD)}} utilizes Gaussian noise to soften the sparse logits of the SNN, better matching with logits of the ANN.
    }\label{fig2}\vspace{-2.5ex}
\end{figure*}

\noindent \textbf{Knowledge Distillation (KD)}~\cite{hinton2014distilling} was initially proposed for compressing artificial neural networks (ANNs) by transferring knowledge from complex teacher models to lightweight student models.
Traditional logits-based KD methods typically minimize the difference in classification probabilities (\textit{i.e.}, logits) between the student and teacher models~\cite{sun2024logit,jin2023multi}.
In addition, feature-based KD methods enhance the performance of the student model by learning the feature representations of the teacher model~\cite{wang2025improving,guo2023class}.
Notably, 
a special class of feature-based KD 
methods~\cite{guo2023class,parchami2024good,zagoruyko2017paying} 
enhances the performance of the student by minimizing the semantic saliency activation maps of both the 
teacher and student models (\textit{e.g.}, Grad-CAM~\cite{selvaraju2017grad}).
Recently, a few works start to apply KD technique to facilitate SNN training~\cite{xu2025bkdsnn,Xu2023kdsnn}, significantly improving the model's performance.
\textit{i.e.}, KDSNN~\cite{Xu2023kdsnn} regularizes consistency of the output features and logits between ANNs and SNNs, while BKDSNN~\cite{xu2025bkdsnn} enhances feature-level matching by further processing the spike features of SNNs with a blurring matrix. 
However, these methods largely overlook the discrepancies of the raw features and logits from teacher ANNs and student SNNs.
Specifically, ANNs generate \textit{continuous floating-point} features, while SNNs generate \textit{discrete spike features}.
The logits of SNNs are also more sparse than that of ANNs.
Thus, we perform KD from the perspective of semantic saliency activation maps, which are more semantically aligned.
But,
prior activation map-based KD methods~\cite{parchami2024good,zagoruyko2017paying} from ANNs cannot be directly applied to SNNs, 
due to the surrogate \textit{gradient estimation errors} in SNNs, which affect the generation of semantic saliency activation maps.
Besides, CATKD~\cite{guo2023class} is only applicable to CNN-based architectures, which limits the application of semantic saliency activation map-based KD in SNNs.
In contrast, we design a semantic saliency activation map-based KD method for the characteristics of SNNs, which can be applied to various architectures.
Moreover, we also design a noise-smoothing distillation method at the logits-level to further enhance the performance of SNNs.
\section{Method}\label{sec:method}

The overall pipeline of our CKDSNN framework is illustrated in Fig.~\ref{fig2}.
Given a pretrained ANN teacher model and a learnable SNN student model, the proposed CKDSNN aims to train the student model by effectively distilling knowledge from the teacher model from two aspects: 
\textbf{1) Saliency-scaled Activation Map Distillation (SAMD)}. The class activation map~\cite{selvaraju2017grad} obtained from the teacher ANN is used to guide the student SNN to fire spikes in salient image regions. During the distillation process, we address the magnitude discrepancy between the two types of activation maps by scaling them into the same range. 
\textbf{2) Noise-smoothed Logits Distillation (NLD).} 
The classification output logits of the student model is smoothed using additional Gaussian noise, making the vanilla sparse logits distribution of student SNN close to the continuous logits distribution of teacher ANN. This facilitates more precise logit-level knowledge distillation. 
We first have a brief introduction to the spiking neuron used in SNN, which leads to the discrepancy on the features and logits of ANN and SNN.
Then, we elaborate on the proposed SAMD and NLD strategies, respectively.

\subsection{Discrepancy Caused by SNN Spiking Neuron}\label{sec:3-1}  

Typical SNNs~\cite{fang2021sew,deng2022tet,meng2023towards,jiang2024tab,deng2024spikingtoken} adopt the integrate-and-fire (IF) neuron as the fundamental unit.
Specifically, the IF neuron first integrates input currents by updating its membrane potential, and then compares it with a pre-set threshold to generate a spike signal, followed by a reset mechanism of the membrane potential.
This process can be formulated as:

\begin{equation}\label{eq1}
    \begin{aligned}
H[t] &= V[t-1] + I[t], \\
S[t] &= \Theta(H[t] - V_{\text{th}}) = 
\begin{cases} 
1, & H[t] \geq V_{\text{th}}, \\ 
0, & H[t] < V_{\text{th}},
\end{cases} \\
V[t] &= H[t](1-S[t]) + V_{\text{reset}} S[t],
\end{aligned}
\end{equation}
where $I[t]$ is the input current at time step $t$ and $V[t{-}1]$ is the membrane potential at previous $t{-}1$ time step. 
$\gamma$ denotes the membrane time constant and
$\Theta(\cdot)$ represents the Heaviside function~\cite{huang2024clif}.
The IF accumulates input currents to update the membrane potential $H[t]$. 
When $H[t]$ exceeds the threshold $V_{\text{th}}$, 
a spike $S[t]$ is generated, 
and the membrane potential is reset to $V_{\text{reset}}$~\cite{meng2023towards}.

In image classification,
the above is attached to each feature encoding block of SNN. 
Specifically, each encoding block contains multiple linear transformation layers, batch normalization layers, and IF spiking neurons, as shown in Fig.~\ref{fig2}(a).
This generates \textit{discrete} spike features \( \boldsymbol{F}^{st} {\in} \mathbb{R}^{T \times C \times H \times W} \) over $T$ time steps, where $C$, $H$, and $W$ denote the number of channels, height, and width of the feature, respectively.
The discrete spike features \( \boldsymbol{F}^{st} \) generated by the IF activation mechanism differ significantly from the \textit{continuous} features \( \boldsymbol{F}^{te} {\in} \mathbb{R}^{C \times H \times W} \) produced by ANN.
This leads to a significant discrepancy in the feature distribution between \( \boldsymbol{F}^{st} \) and \( \boldsymbol{F}^{te} \), resulting in different prediction logits.
Prior KD methods~\cite{Xu2023kdsnn} directly match these two types of features or logits element-wise, ignoring the essential differences in feature representation and distribution, leading to suboptimal distillation results.
We address these issues through carefully designed KD strategies.

\begin{figure}
    \centering
    \includegraphics[width=1\linewidth]{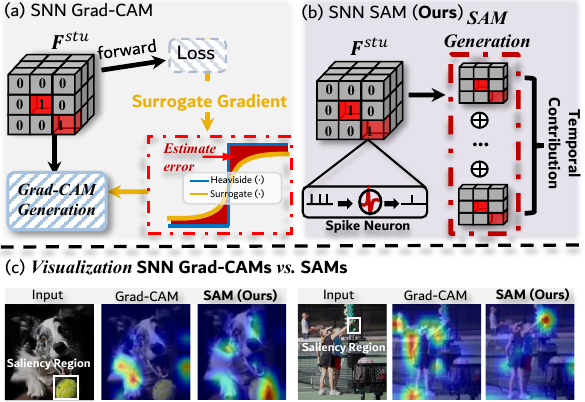} 
    \vspace{-4ex}
    \caption{ 
        Illustration of 
        (a) the main challenge of applying Grad-CAM-like strategies in SNNs is the error caused by surrogate gradients.
        (b) Our SAM directly computes the spike activation rate of SNNs via SAM Generation.
        (c) Visualization of the generated CAMs and SAMs for SNN.
    }
    \label{fig2-sub1}\vspace{-3ex}
\end{figure}

\subsection{Saliency-scaled Activation Map Distillation}\label{sec:3-2}

We propose the {Saliency-scaled Activation Map Distillation} (SAMD) to train SNN rather than distilling the element-wise discrepant features used in existing works~\cite{Xu2023kdsnn,xu2025bkdsnn}.
Specifically, SAMD consists of three steps:

\noindent \textbf{1) Class Activation Map (CAM) Generation}. 
The class activation map is initially used as a visualization tool to enhance the explainability of convolutional neural network~\cite{zhou2016learning}, 
which can highlight the related image region of a specific class.
In our work, we use it as the teacher's knowledge for distillation.
We follow the typical method of Grad-CAM~\cite{selvaraju2017grad} to generate the class activation map.
Specifically,
given an input image \(x\), we first extract the intermediate features using a pretrained teacher ANN model, denoted as $\boldsymbol{F}^{te} {\in} \mathbb{R}^{C\times H\times W}$.
Then, the gradients of $\boldsymbol{F}^{te}$ are calculated according to the forward loss between prediction and the ground truth label $y {\in} \mathbb{R}^{K}$ ($K$ is the total number of classes).
The class activation map \(\boldsymbol{M}^{te} {\in} \mathbb{R}^{H\times W} \) can be obtained by associating \( \boldsymbol{F}^{te}\) with the class label \(y\).
This process can be formulated as follows:

\vspace{-2ex}
\begin{equation}
\begin{aligned}
    \alpha &= \frac{1}{W \cdot H} \sum_{i=1}^{W} \sum_{j=1}^{H} \frac{\partial y}{\partial \boldsymbol{F}_{ i, j}^{te}}, \\
    \boldsymbol{M}^{te} &= \text{ReLU}\left(\sum_{k=1}^{{K}} \alpha_k \boldsymbol{F}_k^{te} \right),
\end{aligned}
\end{equation}
where \(\alpha\), the gradient-based weights from label \(y\), measures each channel's contribution to the target class. Using \(\alpha\) for a weighted sum of the feature map \(\boldsymbol{F}^{te}\) across the channel dimension \(C\) yields the activation map for input image \(x\).

\noindent\textbf{2) Spike Activation Map (SAM) Generation.}
Unlike the gradient-based CAM generation in ANNs, 
the gradient estimation error in SNNs leads to inaccurate activation maps.
As shown in Fig.~\ref{fig2-sub1}(a), 
the gradient-based activation map is not accurate in SNNs, mainly due to the \textit{gradient estimation error}.
Therefore, 
as show in Fig.~\ref{fig2-sub1}(b),
we abandon the gradient-based CAM generation method and design a SAM generation method that directly computes the spike activation map based on the spikes in features.
As shown in Fig.~\ref{fig2-sub1}(c), 
we leverage the characteristics of SNNs, 
where spikes are generated only in salient regions, 
and consider the contribution of spikes at different time steps, 
\textit{i.e.}, spikes at each time step $t$ are accumulated into the SAM.
Specifically,
the intermediate features \( \boldsymbol{F}^{st} {\in} \mathbb{R}^{T\times C \times H \times W} \) can be obtained from a student SNN.
$\boldsymbol{F}^{st}$ indicates the activated spikes over $T$ time steps.
For each time step, the spatial regions are fired with spikes in different channels, where each channel captures key saliency information related to the target class.
Therefore, $\boldsymbol{F}^{st}$ is able to reveal the informative and salient regions.
So we generate the spike activation map $\boldsymbol{M}^{st} {\in} \mathbb{R}^{H \times W}$ by directly averaging $\boldsymbol{F}^{st}$ in the channel and time dimensions, computed as, 
\vspace{-1.5ex}
\begin{equation}\label{eq9}
\boldsymbol{M}^{st} {=} \sum_{t=1}^{T} \sum_{c=1}^{C} \boldsymbol{F}_{t, c}^{st}.
\end{equation} \vspace{-1.0ex}

$\boldsymbol{M}^{st}$ leverages the characteristics of SNN and is able to integrate semantic information across multiple time steps.
Unlike the class activation map generation, this process is computationally efficient and is performed online, allowing later dynamical distillation learning through forward and backward propagation.

\noindent\textbf{3) Saliency-scaled CAM-SAM Distillation.} 
Although both the CAM and SAM highlight the salient regions (semantic aligned), there is a discrepancy between \( \boldsymbol{M}^{te} \) and \( \boldsymbol{M}^{st} \) in the feature magnitude.
This is because that 
\( \boldsymbol{M}^{st} \) is the summarization of discrete SNN binary 0/1 spikes, while \( \boldsymbol{M}^{te} \) is derived by weighting float-point ANN features that range between 0 and 1.
To address this, as shown in Fig.~\ref{fig2}(b), we propose saliency scaling that normalizes \( \boldsymbol{M}^{st} \) and \( \boldsymbol{M}^{te} \) to probability distributions on the same scale using a softmax function: $f(\boldsymbol{M}) {=} {\exp\left(\frac{\boldsymbol{M}}{\mathcal{T}}\right)} / {\sum_{i=1}^{W H} \exp\left(\frac{\boldsymbol{M}}{\mathcal{T}}\right)}$, where \(\mathcal{T}\) is a constant to control the distribution smoothness.

Let $\boldsymbol{P}^{te}$ and $\boldsymbol{P}^{st}$ be the normalized activation map. The CAM-SAM distillation regularizes the consistency of $\boldsymbol{P}^{te}$ and $\boldsymbol{P}^{st}$.
This is achieved by computing the Kullback–Leibler (KL) divergence loss $\mathcal{L}_{\text{SAMD}}$:
\begin{align}\label{eq4}
\mathcal{L}_{\text{SAMD}} &= \mathcal{T}^2 \cdot \text{KL}\left(\boldsymbol{P}^{te} \| \boldsymbol{P}^{st}\right) \nonumber \\
&= \mathcal{T}^2 \cdot \sum_{i=1}^{H} \sum_{j=1}^{W} \boldsymbol{P}^{te}_{ij} \log\left(\frac{\boldsymbol{P}^{te}_{ij}}{\boldsymbol{P}^{st}_{ij}}\right).
\end{align}

\subsection{Noise-smoothed Logits Distillation}\label{sec:3-3} 

In addition to the feature-driven activation map distillation, we consider the target-driven logits distillation, which aims to align the output probability logits of teacher ANN and student SNN.
Although prior work~\cite{Xu2023kdsnn} considered this, they largely overlook the discrepancy between the logits.
This issue still originates from the utilization of different features in logits production where SNNs employ binary spike features that are either 0 or 1, whereas ANNs utilize floating-point features.
Consequently, the logits of SNN are very sparse, while those of ANN are more dense and \textit{continuous}.
We propose the Noise-Smoothed Logits Distillation (NLD) to address this problem.
Thus, we try to soften the logits of student SNN, \( z^{st} {\in} \mathbb{R}^{1\times K} \), reducing the discrepancy from \( z^{te} {\in} \mathbb{R}^{1\times K} \).
Specifically, we add some continuous noise $\epsilon {\in} \mathbb{R}^{1\times K}$ onto \( z^{st}\).
To avoid destroying the original distribution of \( z^{st}\) while maintaining the characteristic of classification logits, we opt to Gaussian noise with the mean and standard deviation of \( z^{st}\):
\begin{equation}\label{noise_add}
    \epsilon \sim \mathcal{N}(\bar{z}^{st}, \sigma(z^{st})^2),
\end{equation}
where $\mathcal{N}$ denotes the Gaussian distribution, $\bar{z}$ and $\sigma(z)$ are the mean and standard variance.

Then, the Gaussian noise $\epsilon$ is fused with $ z^{st}$ using a balance hyper-parameter $\lambda$, computed as,
\begin{equation}\label{lambda_eq}
    z^{soft} = z^{st} +  \lambda \epsilon,
\end{equation}
where $z^{soft}$ is the noise-softened logits of student SNN.
This process introduces randomness through noise, promoting the exploration of a broader decision boundary for classification. 

The softened logits $z^{soft}$ of student SNN and the raw logits $z^{te}$ from the teacher ANN are processed by a softmax function to generate classification probability $y^{soft}$ and $y^{te}$, formulated as $y {=} f(z) {=} {\exp \left(z_i /\tau\right)} / {\sum_{k=1}^K \exp \left(z_k / \tau\right)}$.
Then, logits distillation can be performed by aligning $y^{soft}$ with $y^{te}$ through a KL loss:
\vspace{-1ex}
\begin{equation}\label{eq7}
    \mathcal{L}_{\text{NLD}} = \tau^2 \cdot \text{KL}\left(y^{te} \| y^{soft}\right).
\end{equation}

\subsection{Overall Training Loss} 

Given the student SNN's prediction \(y^{st}\) and the ground truth label \(y\), we can obtain the standard cross-entropy loss \(\mathcal{L}_{\text{CE}}\).
Then, the total objective for model optimization $ \mathcal{L}_{\text{total}}$ is calculated by summarizing the above three losses:
\vspace{-1ex}
\begin{equation}\label{eq16}
    \mathcal{L}_{\text{total}} = \mathcal{L}_{\text{CE}} + \beta \mathcal{L}_{\text{SAMD}} + \gamma \mathcal{L}_{\text{NLD}},
\end{equation}
where $\beta$ and $\gamma$ are hyper-parameters to balance the two distillation losses.

\section{Experiments}
\subsection{Experimental Setups} 

\noindent\textbf{Datasets.} 
We evaluate our method on three widely used 
image classification datasets and a neuromorphic dataset in this research area: 
CIFAR-10/100~\cite{krizhevsky2009learning}, ImageNet-1K~\cite{deng2009imagenet} and CIFAR10-DVS~\cite{li2017cifar10}.
The details of these datasets are provided in the supplementary materials.

\noindent\textbf{Model Configuration.}
To ensure fair comparison, we determine the configurations of teacher ANN and student SNN models based on prior studies~\cite{guo2024enof,xu2025bkdsnn}.
Specifically, 
for the CIFAR-10/100 datasets, we use the ResNet-19/20 versions provided by TET~\cite{deng2022tet} and QCFS~\cite{li2023seenn}, with their corresponding ANN versions as the teacher models; and we also test a ViT-based architecture, using ViT-S as the teacher ANN and Spikformer-4-384~\cite{zhou2023spikformer} as the student model.
For the ImageNet-1K dataset, we use the ResNet-18 and ResNet-34 pre-trained on ImageNet-1K as teacher models, while SEW-ResNet~\cite{fang2021sew} serves as the student model.
For the CIFAR10-DVS dataset, we use ResNet-19 as the student, with its ANN architecture serving as the teacher, trained in the same way as EnOF~\cite{guo2024enof}.

\noindent \textbf{Implementation Details.}
To ensure consistency with prior studies, our implementation on CIFAR-10/100 and ImageNet strictly follows the established distillation architecture~\cite{guo2024enof,xu2025bkdsnn}. 
Specifically, we set the hyperparameters as follows: \(\mathcal{T}\) in Eq.~\ref{eq4} and \(\tau\) in Eq.~\ref{eq7} to \(2.0\).
\(\lambda\) in Eq.~\ref{lambda_eq} to \(0.1\), \(\beta\) and \(\gamma\) in Eq.~\ref{eq16} to \(1.0\).
The Integrate-and-Fire (IF) neuron settings align with prior works~\cite{fang2021sew}, and all other training configurations, including batch size, learning rate, and optimizer, remain consistent with those in~\cite{xu2025bkdsnn}.
In addition, we conduct sensitivity analysis of all hyper-parameters in the supplementary materials.

\noindent \textbf{Platform.}
All experiments are conducted on a server platform equipped with 32 cores Intel Xeon Platinum 8352V CPU with 2.10GHz and 8-way NVIDIA GPUs. We use
SpikingJelly~\cite{fang2023spikingjelly} to simulate the IF~\cite{tal1997computing} spiking neurons.

{
\renewcommand{\arraystretch}{0.7}
\begin{table*}
\centering
\resizebox{0.98\textwidth}{!}{
\begin{tabular}{lcc|cc|cc|cc}
  \toprule
   \multirow{2}{*}{Methods}  & \multirow{2}{*}{Venue} & Time  & \multicolumn{2}{c|}{ResNet20} & \multicolumn{2}{c|}{ResNet19} &  \multicolumn{2}{c}{Spikformer-4-384} \\ 
  &   & step  &  CIFAR10 & CIFAR100 &  CIFAR10 & CIFAR100 & CIFAR10 & CIFAR100  \\ \midrule
  \multicolumn{9}{c}{\textit{ANN-to-SNN}} 
  \\ \midrule
  QCFS~\cite{bu2022optimal} & ICLR'22 & 64  & 92.35 & 55.37 & - & - & - & - \\  \midrule
   \multicolumn{9}{c}{\textit{Direct Training}} 
  \\ \midrule
SEW-R~\cite{fang2021sew} & NIPS'21 & 4 & 89.07 & 60.16 & 93.24 & 70.84 & - & -  \\
STBP~\cite{wu2019direct} & AAAI'21 & 4 & - & - & 92.92 & - & - & -  \\
TET~\cite{deng2022tet} & ICLR'22 & 4 & - & - & 94.44 &  74.47 & - & - \\
SLTT~\cite{meng2023towards} & ICCV'23 & 4 & - & - & 94.56 & 74.67 & - & -  \\
Spikformer~\cite{zhou2023spikformer} & ICLR'22 & 4 & - & - & - & - & 95.93 & 79.65  \\ \midrule
\multicolumn{9}{c}{\textit{SNN-KD}} 
  \\ \midrule
KDSNN~\cite{Xu2023kdsnn} & CVPR'23 & 4 &  89.03 & 60.18 & 94.36 & 74.08 & 95.88 & 80.33  \\ 
 
BKDSNN~\cite{xu2025bkdsnn} & ECCV'24 & 4 & 89.29 & 60.92 & 94.64 & 74.95 & 96.06  & 81.26  \\ 
\multirow{2}{*}{EnOFSNN~\cite{guo2024enof}} & \multirow{2}{*}{NIPS'24} & 1 
       & 92.66 & 70.38 & 95.37 & 77.08 & - & -  \\ 
 & & 2 & 93.86 & 71.55 & 96.19 & 82.43 & - & -  \\
 \rowcolor{gray!20} \textbf{CKDSNN (Ours)} &  - & 1 & \textbf{92.85} & \textbf{72.45} & \textbf{96.11} & \textbf{79.11} & \textbf{96.93} & \textbf{83.07}  \\  
 \rowcolor{gray!20} \textbf{CKDSNN (Ours)} &  - & 2 & \textbf{93.53} & \textbf{73.67} & \textbf{97.13} & \textbf{83.21} & \textbf{96.98} & \textbf{84.53}  \\ 
  \rowcolor{gray!20} \textbf{CKDSNN (Ours)} &  - & 4 & \textbf{94.78} & \textbf{73.88} & \textbf{97.81} & \textbf{83.88} & \textbf{97.54} & \textbf{84.88}  \\ 
 \bottomrule
\end{tabular}
} \vspace{-0.8ex}
\caption{\label{tab1} The comparison of Acc$\uparrow$ (\%) with previous works on CIFAR-10/100 datasets. The best results are \textbf{bolded}.}  \
\vspace{-4.5ex}
\end{table*}
}

\subsection{Comparison with State-of-the-Arts}

We compare our {CKDSNN} with three types of SNN training approaches to evaluate its effectiveness:
1) \textit{ANN-to-SNN}: Conversion of a pre-trained ANN into SNN. 
2) \textit{Direct Training}: Direct training of SNN from scratch. 
3) \textit{SNN-KD}: Training SNN using knowledge distillation 
with the aid of ANN.

\noindent \textbf{Results on CIFAR-10/100.} 
The comparison results between our CKDSNN and previous methods are shown in Tab.~\ref{tab1}.
We significantly outperform existing methods across different architectures.
For example, in the ResNet-19, when the time step is set to 1, CKDSNN achieves an accuracy improvement of 0.74\% on the CIFAR-10 dataset and 1.03\% on the CIFAR-100 dataset compared to the current most competitive knowledge distillation method, EnOF~\cite{guo2024enof}, reaching new SOTA accuracies of 96.11\% and 78.15\%, respectively.
Moreover, notably, as the time step increases, CKDSNN continues to significantly outperform prior works.

\begin{table}[t]
\centering
\resizebox{0.48\textwidth}{!}{
\begin{tabular}{lcc|cc}
  \toprule
   \multirow{2}{*}{\Huge Methods}  & \multirow{2}{*}{ \Huge Venue} & \Huge Time  &  \multirow{2}{*}{\Huge ResNet18} & \multirow{2}{*}{\Huge ResNet34} \\ 
  &   & \Huge step  &  &  \\ \midrule
  \multicolumn{5}{c}{\Huge \textit{ANN-to-SNN}} 
  \\ \midrule
  \Huge QCFS~\cite{bu2022optimal} & \Huge  ICLR'22 &   \Huge 64  & - &   \Huge 72.35 \\  \midrule
   \multicolumn{5}{c}{  \Huge  \textit{Direct Training}} 
  \\ \midrule
  \Huge SEW-R~\cite{fang2021sew} &   \Huge    \Huge  NIPS'21 &   \Huge 4 &   \Huge 63.18 &   \Huge 67.04  \\
  \Huge RMP-Loss~\cite{guo2023rmp} & {  \Huge  ICCV'23} &   \Huge 4 &   \Huge 63.14 &   \Huge 64.71 \\
  \Huge MBPN~\cite{guo2023mbpn} & {   \Huge  ICCV'23} &    \Huge  4 &   \Huge 63.03 &   \Huge 65.17 \\ \midrule
   \multicolumn{5}{c}{  \Huge \textit{SNN-KD}} 
  \\ \midrule
  \Huge KDSNN~\cite{Xu2023kdsnn} &   \Huge  CVPR'23 &   \Huge 4 &   \Huge 63.42 &   \Huge 67.18  \\
  \Huge EnOFSNN~\cite{guo2024enof} &   \Huge  NIPS'24 &   \Huge 4 &   \Huge 65.31 &   \Huge 67.40 \\
  \Huge BKDSNN~\cite{xu2025bkdsnn} &    \Huge  ECCV'24 &   \Huge 4 &   \Huge 65.60 &   \Huge 71.24 \\ 
  \rowcolor{gray!20}   \Huge \textbf{CKDSNN (Ours)} & - &   \Huge 4 &   \Huge \textbf{66.92} &   \Huge \textbf{73.05} \\ 
\bottomrule 
\end{tabular}
} 
\vspace{-0.5ex}
\caption{\label{tab2-in1k}
The comparison of Top-1 Acc$\uparrow$ (\%) with previous works on ImageNet-1K dataset. The best results are \textbf{bolded}.
}
\vspace{-1.0ex}
\end{table}

\noindent \textbf{Results on ImageNet-1K.} The comparison results are presented in Tab.~\ref{tab2-in1k}.
The proposed CKDSNN continues to outperform previous SOTA methods using various teacher models, including ResNet-18, ResNet-34, and ResNet-50. Specifically, compared to the prior SOTA methods BKDSNN~\cite{xu2025bkdsnn}, our CKDSNN improves the top-1 accuracy by 1.32\%, 1.81\%, and 1.52\% using the three types of network architectures, respectively.
These results indicate that CKDSNN is effective on the large-scale dataset.

\noindent \textbf{Results on CIFAR10-DVS.}
The neuromorphic dataset comparison results are shown in Tab.~\ref{tab3_dvs}.
Our CKDSNN also outperforms existing methods on the neuromorphic dataset.
For example, under the same architecture and time step settings, CKDSNN achieves an accuracy improvement of 1.05\% compared to the  most competitive EnOFSNN~\cite{guo2024enof}.

{
\renewcommand{\arraystretch}{0.5}
\begin{table}
    \centering
    \includegraphics[width=0.48\textwidth]{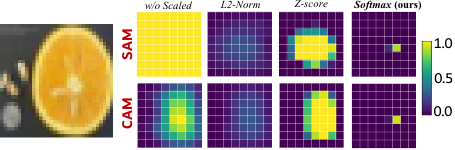} \vspace{-0.3ex}
    \centering
    \resizebox{\linewidth}{!}{
    \begin{tabular}{c|c|c|c|c}
    \toprule
     & \textit{w/o} Scaled & L2-norm  & Z-score  & \cellcolor{gray!20} \textbf{Softmax} \\  \midrule
    Acc$\uparrow$ (\%) & 75.56 & 76.48  & 74.78 & \cellcolor{gray!20} \textbf{79.11} \\ 
    \bottomrule
    \end{tabular}}
    \vspace{-0.5ex}
    \caption{\label{i_s_kl} Ablation study on the saliency-scaling manners in SAMD.}\vspace{-3.5ex}
\end{table}
}

\subsection{Ablation Study} 
We conduct additional ablation experiments to analyze the effectiveness of our proposed strategies.
Unless otherwise specified, all experiments are validated based on the \textbf{ResNet-19} architecture and the \textbf{CIFAR-100} dataset.

\begin{figure}
    \centering
    \includegraphics[width=1\linewidth]{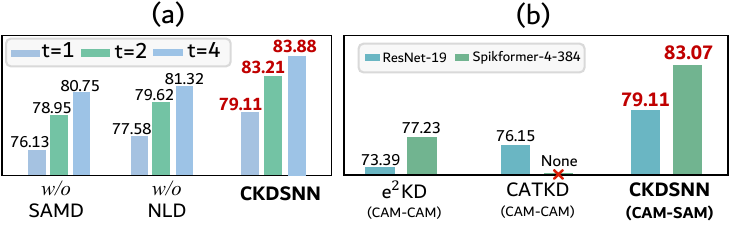}\vspace{-2ex}
    \caption{
    \label{fig-ab1} 
    Ablation study (a) effectiveness of CKDSNN's core strategy. 
    (b) Comparison of SAM-CAM in SAMD with previous activation map-based ANN KD methods at t=1.
    }\vspace{-3.5ex}
\end{figure}

\noindent \textbf{Effectiveness of our core KD strategies.}
Our method primarily consists of the saliency-scaled activation map distillation (SAMD)
and noise-smoothed logits distillation (NLD).
As presented in Fig.~\ref{fig-ab1} (a), without using either SAMD and NLD, the model's performance significantly decreases across various experimental setups.
This indicates the effectiveness and necessity of each proposed KD strategy.

\begin{table} 
\centering
\resizebox{0.48\textwidth}{!}{
\begin{tabular}{lccc|c}
  \toprule
   \multirow{2}{*}{ \LARGE  Methods} & \multirow{2}{*}{\LARGE Venue} & \multirow{2}{*}{\LARGE Arch.} & \LARGE Time  & \multirow{2}{*}{ \LARGE  Acc$\uparrow$(\%)} \\ 
   & & & \LARGE step &  \\ \midrule

 \LARGE STBP~\cite{wu2019direct} & \LARGE  NIPS'21 &  \LARGE ResNet19 &  \LARGE 4 & \LARGE 67.80  \\
 \LARGE SEW-R~\cite{fang2021sew} & \LARGE  NIPS'21 &  \LARGE WideNet &  \LARGE 16 & \LARGE  74.40 \\
 \LARGE KDSNN~\cite{Xu2023kdsnn} & \LARGE  CVPR'23 &  \LARGE ResNet20 &  \LARGE 10 & \LARGE  78.31 \\
 \LARGE BKDSNN~\cite{xu2025bkdsnn} & \LARGE  ECCV'24 &  \LARGE ResNet20 &  \LARGE 10 &  \LARGE  79.53 \\
 \LARGE EnOFSNN~\cite{guo2024enof} & \LARGE  NIPS'24 &  \LARGE ResNet20 &  \LARGE 10 & \LARGE  80.50 \\ 
 \rowcolor{gray!20}  \textbf{ \LARGE  CKDSNN (Ours)} & - &  \LARGE ResNet20 &  \LARGE 10 & \textbf{ \LARGE  81.55} \\ 
\bottomrule
\end{tabular}
}\vspace{-0.3ex}
\caption{\label{tab3_dvs} The comparison with previous works on CIFAR10-DVS dataset. The best results are \textbf{bolded}.}
\vspace{-2ex}
\end{table}

\noindent \textbf{Different saliency-scaling manners in SAMD.}
In SAMD, we use the softmax function to re-scale the class activation map \(\boldsymbol{M}^{te}\) and spike activation map \(\boldsymbol{M}^{st}\) into the same range.
We also try other scaling methods, including \textit{w/o} Scaled, Z-score, and L2-norm.
As reported in Tab.~\ref{i_s_kl}, the softmax scaling strategy exceeds these potential choices by around 2 to 3 points.
When no scaling is applied, although the original value characteristics are preserved, the significant difference in magnitude between the class activation map and spike activation map leads to a significant decrease in saliency alignment effectiveness.
Further analysis in Tab.~\ref{i_s_kl} shows that the softmax scaling strategy effectively normalizes and identifies the most salient regions.
Although other strategies can adjust the value range, they fail to generate a softmax-like probability distribution that normalizes the saliency confidence in activation maps to support alignment.

\noindent \textbf{Importance of CAM-SAM distillation in SAMD.}
The CAM-SAM distillation is one of the core components of SAMD.
As show in Fig.~\ref{fig-ab1} (b), when using ANN-based strategies (\textit{e.g.}, e$^2$KD~\cite{parchami2024good} or CATKD~\cite{guo2023class}) to distill SNNs, the performance is significantly lower than our proposed CAM-SAM distillation.
Specifically, CAM-CAM methods use class Grad-CAM strategies to generate activation maps, but the gradient estimation error leads to performance degradation.
Additionally, CATKD is only applicable to CNN-base architectures. 
In contrast, SAM is designed on the characteristics of SNNs, enabling compatibility with various architectures.

\begin{figure}[t]
\centering
\begin{minipage}{0.45\columnwidth}
\centering
\includegraphics[width=\linewidth]{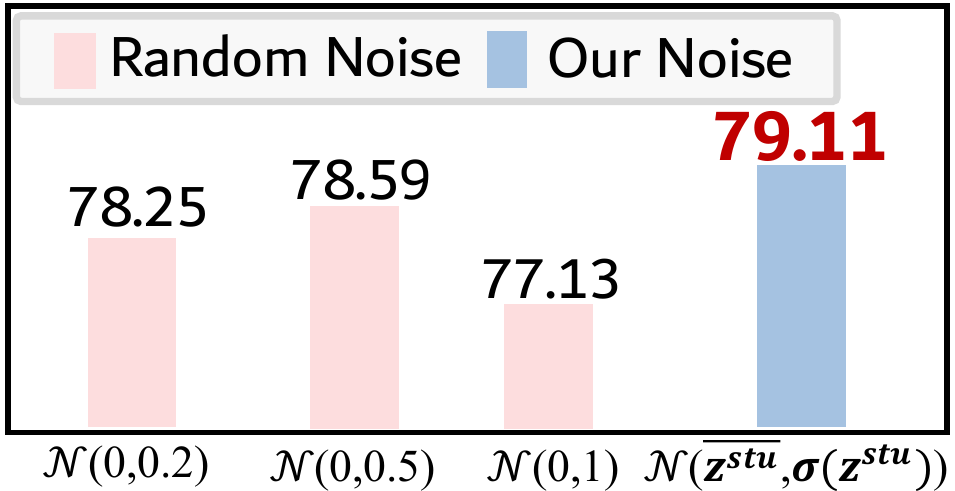}
\vspace{-4ex}
\caption{\label{fig_noise} 
Comparison our adaptive noise strategy with random noise.
}
\end{minipage}%
\vspace{0.5ex}
\begin{minipage}{0.5\columnwidth} 
\centering
\vspace{0.8ex}
\resizebox{\linewidth}{!}{%
\begin{tabular}{lccc}
\toprule
\multirow{2}{*}{\Large Method} & \Large Overhead $\downarrow$  & \Large Acc. $\uparrow$ \\
 &  (min/epoch) & (\%)  \\
\midrule
\Large KDSNN  & \Large 18.20 &  \Large 67.18 \\
\Large BKDSNN & \Large 20.35 & \Large 71.24 \\
\Large CKDSNN & \Large 20.12  &  \Large \textbf{73.05}  \\
\bottomrule
\end{tabular}%
}\vspace{-1.5ex}
\captionof{table}{\label{tab_overhead} CKDSNN's training overhead analysis using ResNet34 on the ImageNet-1k dataset.}
\end{minipage}\vspace{-1.5ex}
\end{figure}

\noindent \textbf{Effect of the logits noise-smoothing in NLD.} 
We compare the effect of random noise and noise-smoothing logits distillation (NLD), as shown in Fig.~\ref{fig_noise}.
Our adaptive noise strategy significantly outperforms random noise 
with a fixed standard deviation: small noise yields poor results, increasing noise improves performance but still lags behind our method, while excessive noise leads to performance degradation.
This validates that our method effectively retains the original distribution characteristics by adaptively adjusting the noise amplitude to
match the SNN logits distribution.

\begin{figure}[t]
    \centering
    \includegraphics[width=1\linewidth]{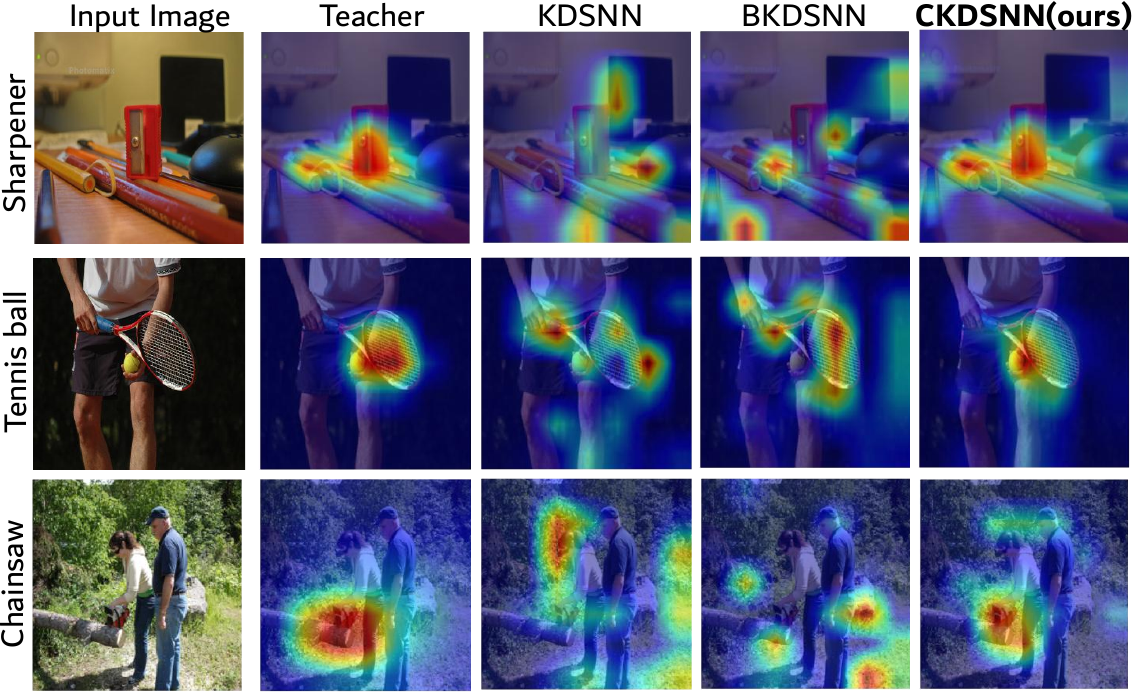}
     \vspace{-4ex}
    \caption{\label{fig3} Visualization of the spike activation maps from different methods. 
    Our {CKDSNN} enables the SNN student to spike in the most salient regions, close to the teacher's knowledge.}\vspace{-3ex}
\end{figure}

\begin{figure}[t]
    \centering
    \includegraphics[width=1\linewidth]{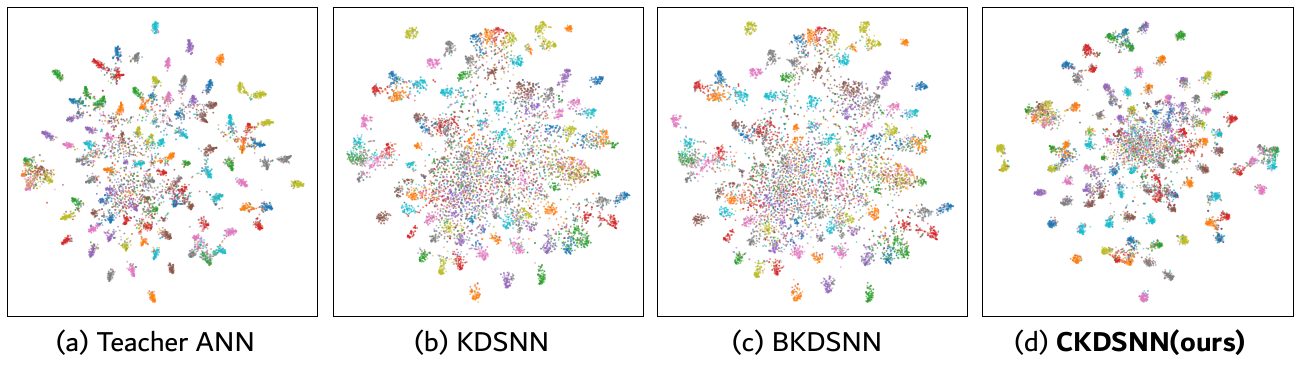}
    \vspace{-3.0ex}
    \caption{\label{fig4} The t-SNE~\cite{van2008visualizing} visualization of feature distributions. (a): teacher ANN's float feature representation. (b)-(d): spike features of student SNNs. Each color denotes an image class.
    }\vspace{-0.5ex}
\end{figure}

\subsection{Qualitative Analysis}

\noindent \textbf{Spiking Activation Maps.}
Fig.~\ref{fig3} visualizes the learned spike activation maps (SAMs) of our method compared to other KD-based approaches. Our SAM aligns more closely with the teacher model's class activation maps (CAMs). For instance, it more accurately captures key regions such as the `sharpener' and `chainsaw', indicating that spikes are precisely emitted on salient areas and the teacher's CAM knowledge is effectively distilled into the student's SAM.

\noindent \textbf{Spiking Features.}  
We extract spiking features from the final layer of our CKDSNN model and visualize them using t-SNE~\cite{van2008visualizing}. 
Image samples from the CIFAR-100 dataset are used.
As shown in Fig.~\ref{fig4}, 
the spike features generated by CKDSNN demonstrate improved separability and discriminability compared to other KD methods and are close to the teacher's feature distribution.
This superiority is largely contributed to the proposed activation map KD, 
which guides the spiking representations to focus on salient regions, resulting in better feature discrimination.

{
\renewcommand{\arraystretch}{0.7} 
\begin{table}[t]
    \centering
    \centering
    \resizebox{\columnwidth}{!}{
    \begin{tabular}{l|ccc|c|c}
    \toprule
    \multirow{2}{*}{\Huge Methods} & {\Huge Fire Rate $\downarrow$} & {\Huge SOPS $\downarrow$ } & {\Huge Power  $\downarrow$ } & {\Huge Acc$\uparrow$}  & {\Huge Time} \\ 
      &\Huge  (\%)  & \Huge (G)  & \Huge (mJ) &  \Huge (\%)  & \Huge Step  \\
    \midrule
    \Huge SEW-\textit{R}~\cite{fang2021sew} &\Huge  18.0 & \Huge 4.14 & \Huge 4.03 &  \Huge 67.04 & \Huge 4 \\ 
    \Huge KDSNN~\cite{Xu2023kdsnn} & \Huge 16.0  & \Huge 4.13 & \Huge 4.01 & \Huge 67.18  & \Huge 4 \\ 
    \Huge  BKDSNN~\cite{xu2025bkdsnn} & \Huge  15.0  & \Huge 4.02 & \Huge 3.98  &  \Huge 71.24 & \Huge  4\\ \midrule
   \multirow{3}{*}{ \textbf{{\Huge  CKDSNN (Ours)}} }   & \textbf{\Huge 8.0}  & \textbf{\Huge 3.73} & \textbf{\Huge 3.61} & \Huge 71.33  & {\Huge 2}  \\
    & {\Huge 10.0} &  { \Huge 3.92 } &  {\Huge 3.88}  &  {\Huge 72.71}  & {\Huge 3} \\
    & \Huge 13.0  & \Huge 4.01 & \Huge  3.96   & \textbf{\Huge 73.05}  & \Huge 4 \\ 
    \bottomrule
    \end{tabular}
}
\vspace{-1.0ex}
\caption{\label{tab_enger} 
Comparison of energy efficiency on the ResNet34 with the ImageNet-1k dataset.
}
\vspace{-2.5ex}
\end{table}
}

\subsection{Efficiency Analysis}

\noindent \textbf{Energy Efficiency.}
We finally evaluate the energy efficiency of our method by calculating the Fire rate, SOPS, and Power consumption following prior methods~\cite{zhou2023spikformer,xu2025bkdsnn,fang2023spikingjelly}.
As shown in Tab.~\ref{tab_enger}, compared with prior KD methods, our CKDSNN not only achieves state-of-the-art performance but offers higher energy efficiency.
For example, our method with the same 4 time steps outperforms BKDSNN~\cite{xu2025bkdsnn} by 1.81\% but using slightly less fire rate and Power.
The energy efficiency of our model can be further improved by reducing the time steps to 2.
In such case, our model still maintains superior performance.
This demonstrates that our method has a better trade-off between the energy-efficiency and accuracy.

\noindent  \textbf{Training overhead.}
We analyze the training overhead of our CKDSNN method.
As reported in Fig.~\ref{tab_overhead}, CKDSNN incurs a training overhead that is higher than KDSNN but lower than BKDSNN, while achieving the best performance.

\section{Conclusion}
\label{sec:conclusion}
This paper takes a closer look at current SNN training methods using knowledge distillation (KD) techniques and finds that the discrepancies of features and logits between teacher ANNs and student SNNs are largely overlooked.
We propose two novel KD strategies. 
The saliency-scaled activation map distillation aligns spike activation map from student SNN with the class activation map from teacher ANN.
The noise-smoothed logits distillation aligns the teacher ANN's classification logits with student SNN's logits softened by Gaussian noise.
In this way, the saliency activation map and the logits from the teacher and student are more semantic- and distribution-consistent, guaranteeing more effective knowledge distillation in SNN training.
Extensive experiments demonstrate the effectiveness and robustness of our method.

\section{Acknowledgements}
This work was supported in part by the National Key R\&D Program of China (Grant No. 2024YFB3311602);
the National Natural Science Foundation of China (Grants 61971178, 62501224, 62272144, and U20A20183); 
the Ordos Science and Technology Major “Open Bidding” Project (Grant No. JBGS-2023-002); 
the State Grid Co., Ltd. Headquarters Technology Project (Grant No. 5500-202140127); 
the Major Project of Anhui Province (Grant No. 202423k09020001); 
the Anhui Provincial Natural Science Foundation for Distinguished Young Scholars (Grant No. 2408085J040); 
the Fundamental Research Funds for the Central Universities (Grants JZ2024HGTG0309, JZ2024AHST0337, JZ2024AKKG0507); the Fundamental Research Funds for the Central Universities (JZ2025HGTA0160,JZ2025HGQA0139)
and the Anhui Provincial Water Conservancy Science and Technology Project (Grant No. slky202501-06).

\bibliography{main}

\newpage

\twocolumn[{
  \centering
  \fontsize{14.5}{17.4}\selectfont 
  \textbf{Supplementary Material}
  \par
  \vspace{2em}
}]

\noindent We provide additional experimental details in the supplementary material, including:
\begin{itemize}
  \item 1) The detailed descriptions of all datasets used in our experiments.
  \item 2) The specific implementation details of the models.
  \item 3) The theoretical analysis of the proposed methods.
  \item 4) The sensitivity analysis results of hyper-parameters.
  \item 5) The additional ablation study results.
  \item 6) The additional visualization results.
\end{itemize}

\section{Datasets}
\noindent \textbf{CIFAR10.}~\cite{krizhevsky2009learning} is a widely used dataset for image classification, consisting of 60,000 32x32 color images in 10 classes, with 6,000 images per class. The dataset is divided into 50,000 training images and 10,000 test images.
In addition, we apply data augmentation techniques including data normalization, 
random horizontal flipping, random cropping, AutoAugment~\cite{cubuk2019autoaugment}, 
and Cutout~\cite{devries2017improved}, consistent with prior works~\cite{guo2024enof}.

\noindent \textbf{CIFAR100.}~\cite{krizhevsky2009learning} is similar to CIFAR10 but contains 100 classes, each with 600 images. The dataset is also split into 50,000 training images and 10,000 test images.
The data augmentation methods are consistent with those used for CIFAR10.

\noindent \textbf{ImageNet.}~\cite{deng2009imagenet} is a large-scale dataset for image classification, containing over 14 million images across 1,000 classes. The dataset is divided into training and validation sets, with 1.2 million training images and 50,000 validation images.

\noindent \textbf{CIFAR10-DVS.}~\cite{li2017cifar10} is a neuromorphic version of the CIFAR-100 dataset, containing 10 classes with 10,000 images. 
We follow the principles of prior works~\cite{wu2019direct} to split the dataset into 9,000 training images and 1,000 test images, resizing the images to 48x48 for model evaluation.
In addition, we adopt the data augmentation strategy from~\cite{guo2022recdis}, which includes random horizontal flipping and random cropping within 5 pixels.

\section{Implementation Details}

In our experiments, we use 8 NVIDIA 3090 GPUs for training on the ImageNet dataset, 
setting the batch size to 64 and the initial learning rate to 0.1. 
We employ the SGD optimizer with momentum set to 0.9 and weight decay of 1e-4, 
along with a Cosine Annealing learning rate scheduler. 
For the CIFAR10 and CIFAR100 datasets, 
we utilize 2 NVIDIA 3090 GPUs with a batch size of 128, 
keeping the other parameters consistent with those used for prior works~\cite{guo2024enof,xu2025bkdsnn}. 
For the CIFAR100-DVS neuromorphic dataset, 
we use a single NVIDIA 3090 GPU and follow the EnoFSNN principles~\cite{guo2024enof} to adjust the first layer channel count of the ANN to 20, allowing it to process all time inputs at once. All models are implemented using PyTorch and PyTorch-Lightning frameworks.

\begin{figure}
    \centering
    \includegraphics[width=1\linewidth]{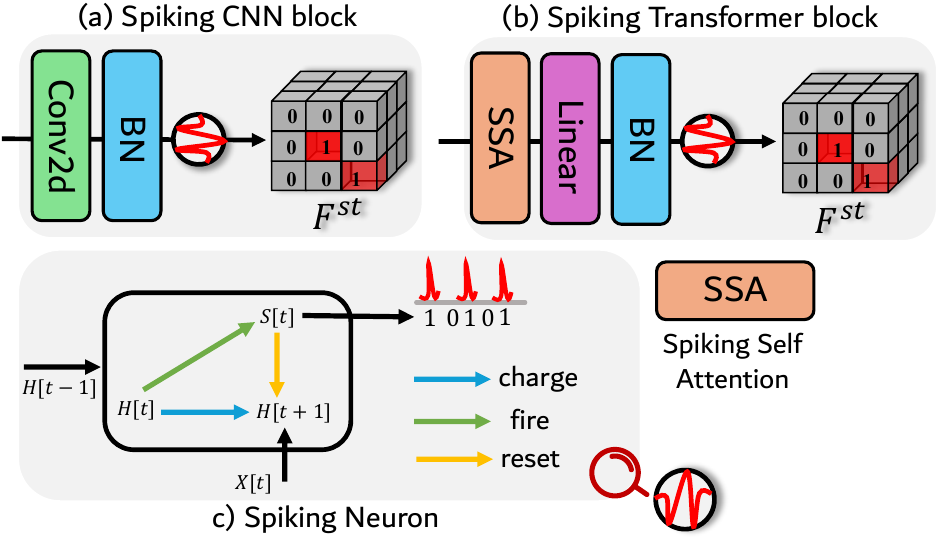}
    \vspace{-3ex}
    \caption{
    Illustration of two typical spiking neural network blocks:
    (a) Spiking CNN block, which consists of Conv2d, BatchNorm, and spiking neurons, configured in architectures like Spiking ResNet or SEW-ResNet to produce discrete features \(\boldsymbol{F}^{st}\).
    (b) Spiking Transformer block, which consists of spiking self-attention, linear transformation layers, and BatchNorm, configured in architectures like Spikformer to produce discrete features \(\boldsymbol{F}^{st}\).
    (c) The process of spiking neurons emitting spikes, where \(S[t]\) is the output of the spiking neuron and \(\boldsymbol{F}^{st}\) is the discrete feature.
    } \label{fig-s-f} 
    \vspace{-2ex}
\end{figure}

\section{Theoretical Analysis}

\subsection{Proof of SAMD Effectiveness}

\noindent \textbf{The non-differentiability of Heaviside function.}
As shown in Fig.~\ref{fig-s-f}, 
SNN's spiking neuron uses the Heaviside step function to generate spikes \( \boldsymbol{F}^{st} \).
However, the Heaviside function $\Theta(\cdot)$ is non-differentiable at \( x = 0 \), 
and its derivative is given by:
\begin{equation}
  \frac{d\Theta(x)}{dx} = \delta(x),
\end{equation}
where \( \delta(x) \) is the dirac delta function, which is non-zero only at \( x = 0 \).
This means that at \( V[t] = V_{\text{th}} \), 
the gradient is either zero (not providing effective information) 
or infinite (not directly usable for backpropagation).

\noindent \textbf{The introduction of surrogate gradient.}
To avoid the infinite values during backpropagation,
we introduce the Sigmoid function as a surrogate gradient function for the Heaviside function.
The Sigmoid function is defined as:
\begin{equation}
  \sigma(x) = \frac{1}{1 + e^{-kx}},
\end{equation}
where \( k \) is a hyperparameter controlling the steepness of the Sigmoid function 
(as \( k \to \infty \), the Sigmoid approaches the Heaviside function).

In backpropagation,
we assume that the forward pass still uses the Heaviside function \( S[t] {=} \Theta(V[t] - V_{\text{th}}) \),
but during gradient computation, the derivative of the Heaviside function 
is replaced by the derivative of the Sigmoid function:
\begin{equation}
  \frac{d\sigma(x)}{dx} = k \sigma(x) (1 - \sigma(x)),
\end{equation}
where \( x = V[t] - V_{\text{th}} \).
This means that for the output spike \( S[t] \),
the gradient with respect to the membrane potential \( V[t] \) is approximated as:
\begin{equation}
  \frac{\partial S[t]}{\partial V[t]} \approx k \sigma(V[t] - V_{\text{th}}) (1 - \sigma(V[t] - V_{\text{th}})).
\end{equation}

\noindent \textbf{The gradient error estimation of surrogate gradient.}
We assume the loss function is \( L \), and we focus on the gradient of the loss with respect to a weight \( w \), denoted as \( \frac{\partial L}{\partial w} \).
If the Heaviside function were differentiable, its gradient would be \( \delta(V[t] - V_{\text{th}}) \).
The gradient of the loss with respect to the weight \( w \) would be computed using the chain rule:
\begin{equation}
  \frac{\partial L}{\partial w} = \sum_t \frac{\partial L}{\partial S[t]} \cdot \frac{\partial S[t]}{\partial V[t]} \cdot \frac{\partial V[t]}{\partial w},
\end{equation}
the term \( \frac{\partial S[t]}{\partial V[t]} = \delta(V[t] - V_{\text{th}}) \) is the gradient of the Heaviside function, and \( \frac{\partial V[t]}{\partial w} \) depends on the input spikes \( S[t] \) and the membrane potential update.

Because \( \delta(V[t] - V_{\text{th}}) \) is zero when \( V[t] \neq V_{\text{th}} \) and non-zero only at \( V[t] = V_{\text{th}} \), the gradient information is highly sparse and unstable.

The gradient using the Sigmoid surrogate gradient becomes:
\begin{equation}
  \frac{\partial S[t]}{\partial V[t]} \approx k \sigma(V[t] - V_{\text{th}}) (1 - \sigma(V[t] - V_{\text{th}})).
\end{equation}

Thus, the approximate gradient is:
\begin{equation}
  \frac{\partial L}{\partial w}_{\text{approx}} \approx 
  \sum_t \frac{\partial L}{\partial S[t]} \cdot k \sigma(V[t] - V_{\text{th}}) (1 - \sigma(V[t] - V_{\text{th}})) 
  \cdot \frac{\partial V[t]}{\partial w}.
\end{equation}

We assume the gradient in the non-differentiable case is \( \frac{\partial L}{\partial w} \), and the approximate gradient is \( \frac{\partial L}{\partial w}_{\text{approx}} \).
The error between the true gradient and the approximate gradient can be expressed as:
\begin{equation}
  \text{Error} = \left| \frac{\partial L}{\partial w} - \frac{\partial L}{\partial w}_{\text{approx}} \right|.
\end{equation}

The true gradient depends on \( \delta(V[t] - V_{\text{th}}) \),
while the approximate gradient depends on \( k \sigma(V[t] - V_{\text{th}}) (1 - \sigma(V[t] - V_{\text{th}})) \).
The error arises from the differences in shape and magnitude between the two:
\begin{itemize}
  \item When \( V[t] \approx V_{\text{th}} \), the Sigmoid derivative provides a non-zero gradient, while the true gradient is infinite or undefined.
  \item When \( V[t] \gg V_{\text{th}} \) or \( V[t] \ll V_{\text{th}} \), the Sigmoid derivative approaches zero, but its non-zero value may introduce additional noise.
\end{itemize}

\noindent \textbf{The impact of gradient estimation error on Grad-CAM.}
The Grad-CAM activation map is computed as:
\begin{equation}
\begin{aligned}
    \alpha^{s} &= \frac{1}{W \cdot H} \sum_{i=1}^{W} \sum_{j=1}^{H} \frac{\partial y}{\partial \boldsymbol{F}_{ i, j}^{st}}, \\
    \boldsymbol{CAM}^{st} &= \text{ReLU}\left(\sum_{k=1}^{{K}} \alpha^{s}_k \boldsymbol{F}_k^{st} \right),
\end{aligned}
\end{equation}
as shown in Fig.~\ref{fig-s-f}, 
the feature map \( \boldsymbol{F}^{st} \) is typically the spiking sequence \( S[t] \), 
and the gradient \( \frac{\partial y^c}{\partial S[t]} \) depends on the surrogate gradient.
Since the Sigmoid surrogate gradient differs from the true gradient (Dirac delta),
the computed \( \alpha \) deviates from the true contribution, 
leading to a bias in the activation map $\boldsymbol{CAM}^{st}$.

\noindent \textbf{The effectiveness of SAMD.}
We can conclude from the above analysis that the use of surrogate gradient leads to a bias in the Grad-CAM activation map $\boldsymbol{CAM}^{st}$.
Therefore, we redesigned the activation map generation method in SNNs,
redesigned the activation map generation method in SNNs.
We shifted to using the Spiking Activation Map (SAM) instead of the Grad-CAM activation map.
In the generation of SAM, we do not rely on gradient information,
but directly count the number of spikes to generate high-quality activation map.
Our experiments also demonstrate that the SAM generated in this way can effectively capture the key regions of the input image.
The CAM-SAM distillation from the teacher ANN to the student SNN leads to spikes being generated only in the most salient regions, achieving better distillation results.

\subsection{Proof of the NLD Effectiveness}

\textbf{Maximum Entropy Principle} is a probabilistic modeling method in information theory~\cite{jaynes1957information}, which aims to select the probability distribution with maximum entropy under given constraints. The entropy is defined as:
\begin{equation}
  H(q) = -\sum_{k=1}^K q^{(k)} \log q^{(k)},
\end{equation}
where $q^{(k)}$ represents the probability of class $k$ in the probability distribution, and $K$ is the total number of classes. 
The entropy $H(q)$ measures the uncertainty of the distribution; 
a larger entropy indicates a more uniform distribution, 
meaning fewer assumptions about the part of the information not provided given the known information.
According to the maximum entropy principle, given cetrain constraints (\textit{e.g.}, probability normalization), 
we seek a probability distribution $q$ that maximizes its entropy:
\begin{equation}
  \max_q H(q) = -\sum_{k=1}^K q^{(k)} \log q^{(k)}.
\end{equation}
Suppose there are no other constraints, only the probability normalization constraint:
\(\sum_{k=1}^K q^{(k)} = 1\).

To solve the constrained optimization problem, 
we use the method of Lagrange multipliers. 
We construct the Lagrangian function, combining the entropy maximization problem 
(\textit{i.e.}, minimizing the negative entropy $-\sum_{k=1}^K q^{(k)} \log q^{(k)}$) with the constraints:
\[
\mathcal{L}(q, \lambda) = -\sum_{k=1}^K q^{(k)} \log q^{(k)} + \lambda \left( \sum_{k=1}^K q^{(k)} - 1 \right),
\]
the first term $-\sum_{k=1}^K q^{(k)} \log q^{(k)}$ is the negative value of the entropy $H(q)$, 
and maximizing $H(q)$ is equivalent to minimizing $-\sum_{k=1}^K q^{(k)}  \log q^{(k)}$.
the second term $\lambda \left( \sum_{k=1}^K q^{(k)} - 1 \right)$ introduces the 
Lagrange multiplier $\lambda$ to handle the probability normalization constraint $\sum_{k=1}^K q^{(k)} = 1$.
According to the method of Lagrange multipliers, 
we can derive that the \textbf{softmax} is the unique solution to the maximum entropy distribution, but its specific form depends on the constraints~\cite{sun2024logit}.
When only the probability normalization constraint is present, the solution is a uniform distribution:
\[
q^{(k)} = \frac{1}{K}, \quad \forall k,
\]
it indicates that, in the absence of additional information, the \textit{uniform distribution is the distribution with maximum entropy}.

\noindent \textbf{The influence of constraints in KD.}
In the knowledge distillation (KD), the optimization of the student model can be viewed as a constrained optimization problem, similar to the maximum entropy derivation above. (1) normalization constraint:
\(\sum_{k=1}^K q(z^{st})^{(k)} = 1\), 
(2) expectation constraint:
\(
\sum_{k=1}^K z^{st (k)} q(z^{st})^{(k)} = \sum_{k=1}^K z^{st (k)} q(z^{te})^{(k)}
\),
the Lagrangian function is constructed using the method of Lagrange multipliers:
\vspace{-2ex}
\begin{align}
\mathcal{L}(q, \lambda, \beta) 
&= -\sum_{k=1}^K q(z^{st})^{(k)} \log q(z^{st})^{(k)} \nonumber \\
&\quad + \lambda \left( \sum_{k=1}^K q(z^{st})^{(k)} - 1 \right) \nonumber \\
&\quad + \beta \left( \sum_{k=1}^K z^{st (k)} q(z^{st})^{(k)} - \sum_{k=1}^K z^{st (k)} q(z^{te})^{(k)} \right).
\end{align}
By taking the derivative with respect to \( q(z^{st})^{(k)} \) and setting it to zero, we obtain:
\begin{equation}
  q(z^{st})^{(k)} \propto \exp(\beta z^{st (k)}).
\end{equation}
The normalized form leads to the softmax distribution.
This form indicates that the distribution of the student model 
is constrained to align with the statistical properties of the teacher model's output while maintaining high entropy as much as possible.

\noindent \textbf{Drawbacks of SNNs' probability distribution in KD.}
In SNN-KD, the output distribution of the student model $y^{st}$ aims to be as uniform as possible while satisfying the constraints (\textit{i.e.}, matching the teacher model $y^{te}$ and the true label $y$).
However, due to the discrete nature of spikes, the logits $y^{st}$ of SNNs have a sparse and sharp value range, leading $y^{st}$ to tend towards a low-entropy distribution (\textit{i.e.}, close to one-hot).
This conflicts with the maximum entropy objective, making it difficult to effectively learn from $y^{te}$.

\noindent \textbf{The noise-smooth effectiveness.}
To address the above issue, we propose the \textbf{noise-smooth logits} strategy, which aims to alleviate the low-entropy distribution problem of SNNs by introducing noise smoothing.
Through adding noise to the logits of SNNs, we can make their distribution smoother, thereby avoiding the problem of low entropy in distillation and enabling effective distillation learning.

\section{Hyperparameter Experiments}

\begin{figure}
    \centering
    \includegraphics[width=1\linewidth]{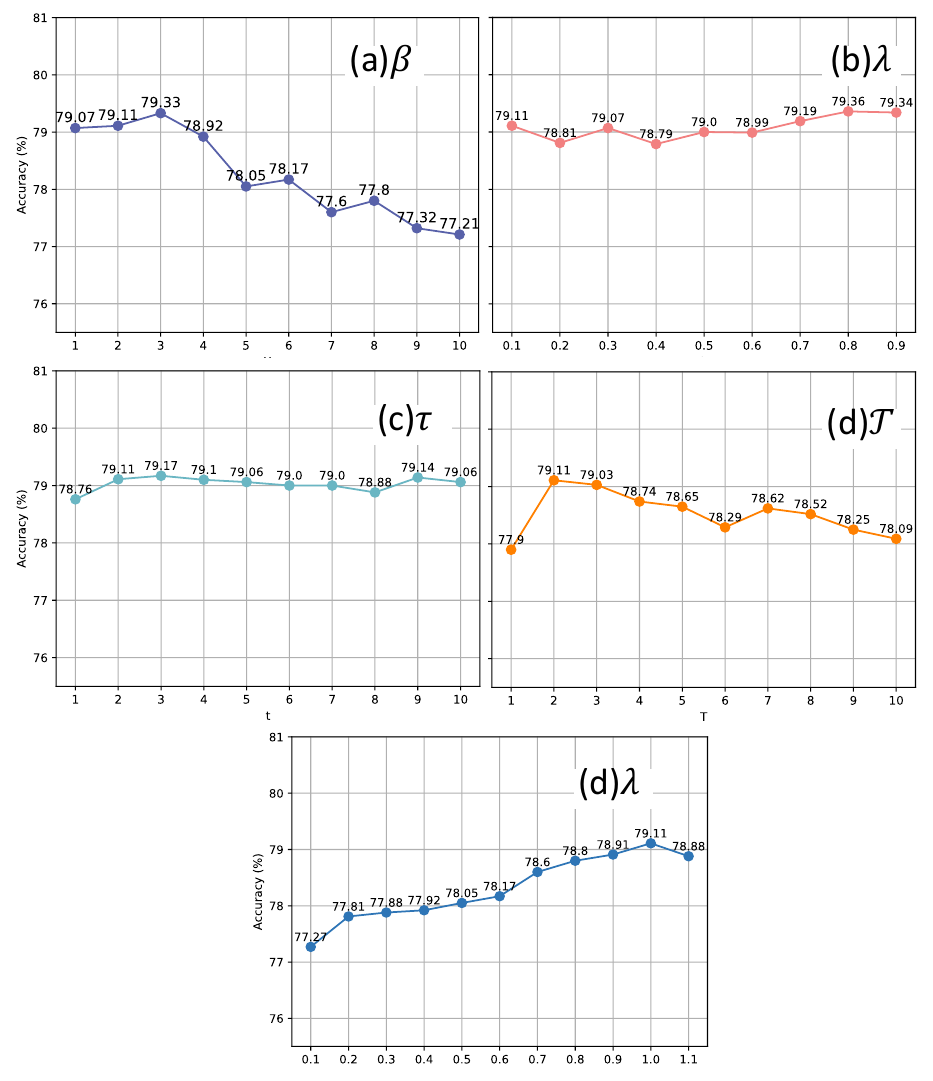}\vspace{-2ex}
    \caption{\label{hyper-all}
    The sensitivity analysis of hyper-parameters on CIFAR100 using ResNet-19, including: $\beta$, $\lambda$, $\tau$, $\mathcal{T}$ and $\gamma$.
    }\vspace{-1ex}
\end{figure}

We conducted experiments on all hyperparameters used in our experiments, and the results are shown in Figure~\ref{hyper-all}.

{
\renewcommand{\arraystretch}{0.6}
\begin{table}[t]
\centering
\resizebox{0.98\linewidth}{!}{
\begin{tabular}{c|>{\centering}p{0.8cm}|>{\centering}p{0.8cm}|>{\centering}p{0.8cm}|c|c}
\toprule
 {Stage} & {1} & {2} & {3}   & {\textbf{4}}  & {all}\\  \midrule 
 Acc. & 77.01 & 77.04 & 78.25 & \textbf{79.11} & 78.93  \\ 
\bottomrule
\end{tabular}
}
\vspace{-2ex}
\caption{\label{i_s_L} Ablation study on the position (\textit{i.e.}, stage of the ANN/SNN models) of applying SAMD. The ResNet-19 is used as the teacher and student models, respectively. Experiments are conducted on the CIFAR-100 dataset.}
\vspace{-1ex}
\end{table}
}

\begin{figure*}
    \centering
    \includegraphics[width=1\linewidth]{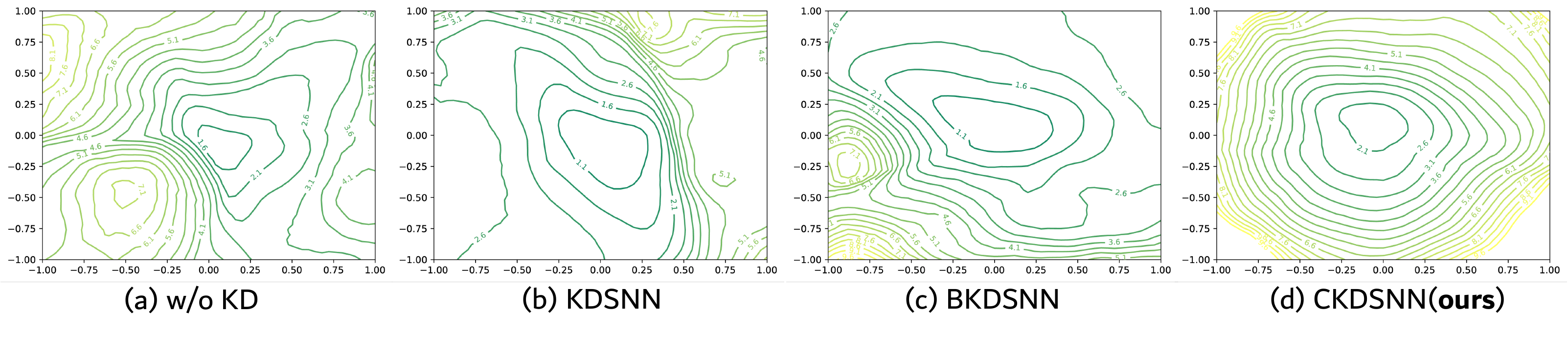}
    \vspace{-5ex}
    \caption{ Visualization of 2D loss landscape produced by different methods. Compared to without using KD or using prior KD methods (\textit{i.e.}, KDSNN~\cite{Xu2023kdsnn} and BKDSNN~\cite{xu2025bkdsnn}), student model trained with our {CKDSNN} has flatter loss landscape with fewer saddle points, leading to a smoother optimization path toward global minima.}\label{fig13}\vspace{-2ex}
\end{figure*}

\begin{figure}
    \centering
    \includegraphics[width=0.995\linewidth]{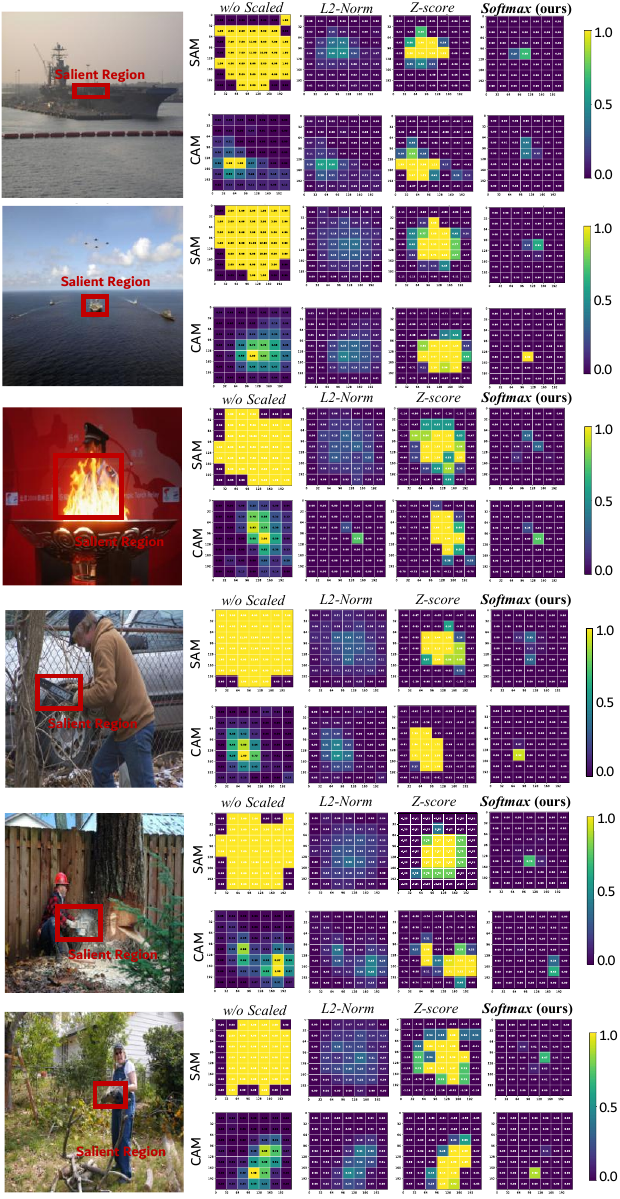}\vspace{-1ex}
    \caption{\label{fig10}
    Comparison of different saliency-scaling manners in SAMD on ImageNet-1k dataset.
    We also observe that the softmax scaling strategy effectively identifies the most salient regions.
    }\vspace{-3ex}
\end{figure}

\section{Others Ablation Studies}

\noindent\textbf{{Applying SAMD in different stages.}}
Teacher and student architectures, \textit{i.e.}, the ResNets and Transformer-based models, usually have four stages.
We conduct experiments to explore the effects of applying our SAMD strategy in different stages.
Tab.~\ref{i_s_L} reports the best performance when using SAMD at the final stage.
This aligns with the convention of prior methods~\cite{xu2025bkdsnn,kim2021visual} to extract the class activation map (CAM) and spike activation map (SAM) at the final stage.
The results show that CAM and SAM from deeper layers contain more precise semantic information related to the target class, resulting in a more effective distillation.

\section{More Visualization Evidence}


\noindent \textbf{Saliency-scaled Visualization.}
We further conducted visualizations of different saliency scaling methods on ImageNet-1k,
and the results are similar to those on CIFAR datasets.
As shown in Figure~\ref{fig10}, the softmax scaling method effectively normalizes the saliency maps to the most salient regions, outperforming other normalization methods.

\noindent \textbf{Loss Landscape Visualization.}
In Fig.~\ref{fig13}, we visualize the loss landscape of different KD methods.
As can be seen,
our method CKDSNN has a much flatter loss landscape compared with both the model trained without (\textit{w/o}) KD and prior KD methods such as KDSNN~\cite{Xu2023kdsnn} and BKDSNN~\cite{xu2025bkdsnn}.
This indicates that our model can avoid the sharp convergence and gradient oscillation issues typically encountered around local minima during the training process, thereby guaranteeing more effective model learning.
This should be thanks to the proposed noise-smoothed logits distillation strategy, supporting the student SNN model explore a smoother decision space to converge to a flatter local minima.

\begin{figure}
    \centering
    \includegraphics[width=0.99\linewidth]{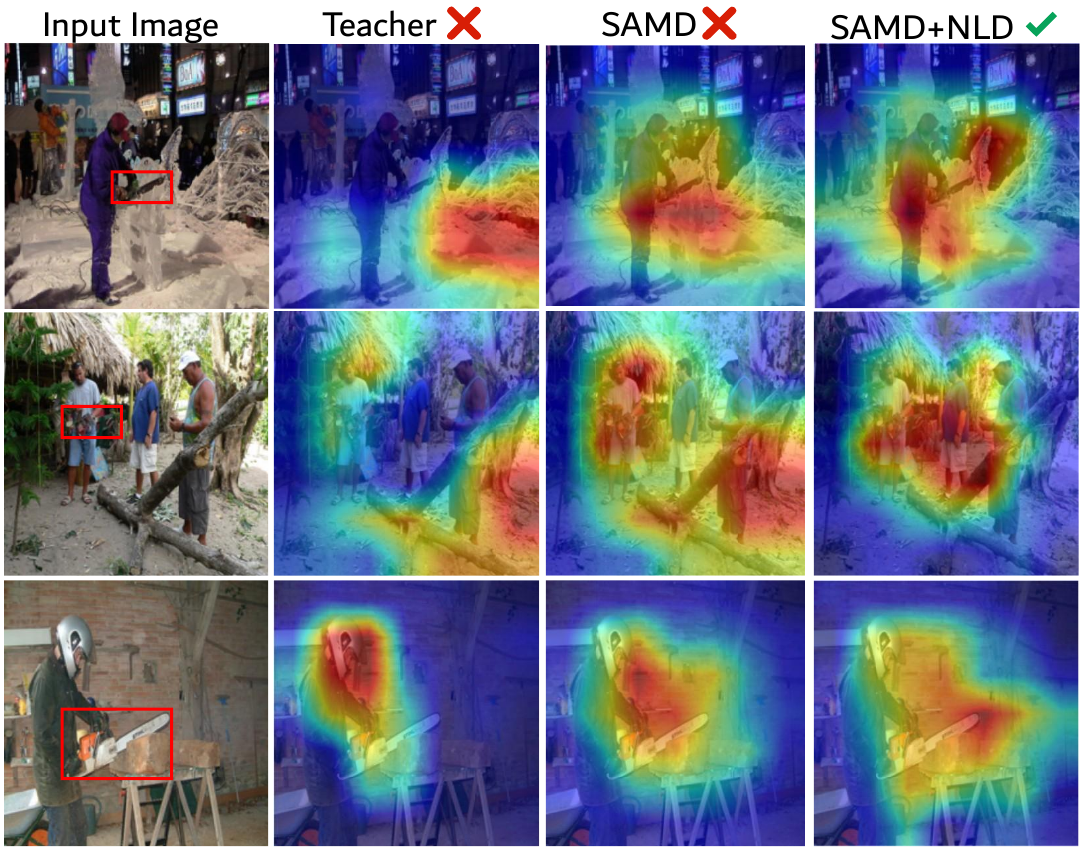} \vspace{-2ex}
    \caption{\label{fig-t-error} 
    The NLD strategy is introduced when using SAMD to improve localization accuracy in cases where the teacher's CAM contains errors.
    } \vspace{-3ex}
\end{figure}

\noindent \textbf{The impact of teacher CAM error on student SAM.}
Another insightful issue is that the error in the teacher ANN's CAM can affect the student model's SAM.
We analyze this issue and find that when only using SAMD, the error in the teacher ANN's CAM leads to some errors in the student SNN's SAM localization.
However, when NLD is added, the student SNN's SAM localization becomes more accurate.
The Fig.~\ref{fig-t-error} illustrates this phenomenon.

\end{document}